\documentclass[sigconf]{acmart}
\usepackage[utf8]{inputenc} 
\usepackage[T1]{fontenc}    
\usepackage{hyperref}       
\usepackage{url}            
\usepackage{amsfonts}       
\usepackage{nicefrac}       
\usepackage{microtype}      
\usepackage{xcolor}         
\usepackage{subfigure}
\usepackage{multirow}
\usepackage{wrapfig}
\usepackage{algorithm}
\usepackage{balance}
\usepackage{algorithmic}

\usepackage{natbib}
\usepackage{mathtools}
\usepackage{amsthm}
\usepackage{bm}
\newtheorem{definition}{Definition}
\newtheorem{thm}{Theorem}

\usepackage{tablefootnote}

\AtBeginDocument{%
 }

\copyrightyear{2024}
\acmYear{2024}
\setcopyright{None}
\acmConference[KDD '24]{Proceedings of the 30th ACM SIGKDD Conference on Knowledge Discovery and Data Mining}{August 25--29, 2024}{Barcelona, Spain}
\acmBooktitle{Proceedings of the 30th ACM SIGKDD Conference on Knowledge Discovery and Data Mining (KDD '24), August 25--29, 2024, Barcelona, Spain}\acmDOI{10.1145/3637528.3671732}
\acmISBN{979-8-4007-0490-1/24/08}

\makeatletter
\gdef\@copyrightpermission{
  \begin{minipage}{0.3\columnwidth}
    \href{https://creativecommons.org/licenses/by/4.0/}{\includegraphics[width=0.90\textwidth]{figures/4ACM-CC-by-88x31.eps}}
  \end{minipage}\hfill
  \begin{minipage}{0.7\columnwidth}
   \href{https://creativecommons.org/licenses/by/4.0/}{This work is licensed under a Creative Commons
Attribution International 4.0 License.}
  \end{minipage}
  \vspace{5pt}
}
\makeatother

\begin{document}

\title{Topology-aware Embedding Memory for Continual Learning on Expanding Networks}

\author{Xikun Zhang}
\email{xzha0505@uni.sydney.edu.au}
\affiliation{%
  \institution{The University of Sydney}
  \city{Sydney}
  \state{NSW}
  \country{Australia}
}

\author{Dongjin Song}
\email{dongjin.song@uconn.edu}
\affiliation{%
  \institution{University of Connecticut}
  \city{Storrs}
  \state{CT}
  \country{USA}
}

\author{Yixin Chen}
\email{chen@cse.wustl.edu}
\affiliation{%
  \institution{Washington University in St. Louis}
  \city{St. Louis}
  \state{MO}
  \country{USA}
}

\author{Dacheng Tao}
\email{dacheng.tao@gmail.com}
\affiliation{%
  \institution{The University of Sydney}
  \city{Sydney}
  \state{NSW}
  \country{Australia}
}

\renewcommand{\shortauthors}{Xikun Zhang, Dongjin Song, Yixin Chen, and Dacheng Tao}

\begin{abstract}
  Memory replay based techniques have shown great success for continual learning with incrementally accumulated Euclidean data. Directly applying them to continually expanding networks, however, leads to the potential memory explosion problem due to the need to buffer representative nodes and their associated topological neighborhood structures. To this end, we systematically analyze the key challenges in the memory explosion problem, and present a general framework, \textit{i.e.}, Parameter Decoupled Graph Neural Networks (PDGNNs) with Topology-aware Embedding Memory (TEM), to tackle this issue. The proposed framework not only reduces the memory space complexity from $\mathcal{O}(nd^L)$ to $\mathcal{O}(n)$~\footnote{$n$: memory budget, $d$: average node degree, $L$: the radius of the GNN receptive field}, but also fully utilizes the topological information for memory replay. Specifically, PDGNNs decouple trainable parameters from the computation ego-subnetwork via \textit{Topology-aware Embeddings} (TEs), which compress ego-subnetworks into compact vectors (\textit{i.e.}, TEs) to reduce the memory consumption. Based on this framework, we discover a unique \textit{pseudo-training effect} in continual learning on expanding networks and this effect motivates us to develop a novel \textit{coverage maximization sampling} strategy that can enhance the performance with a tight memory budget. Thorough empirical studies demonstrate that, by tackling the memory explosion problem and incorporating topological information into memory replay,  PDGNNs with TEM significantly outperform state-of-the-art techniques, especially in the challenging class-incremental setting. 
\end{abstract}




\maketitle

\section{Introduction}

\begin{figure}
    \centering
    \includegraphics[width=0.45\textwidth]{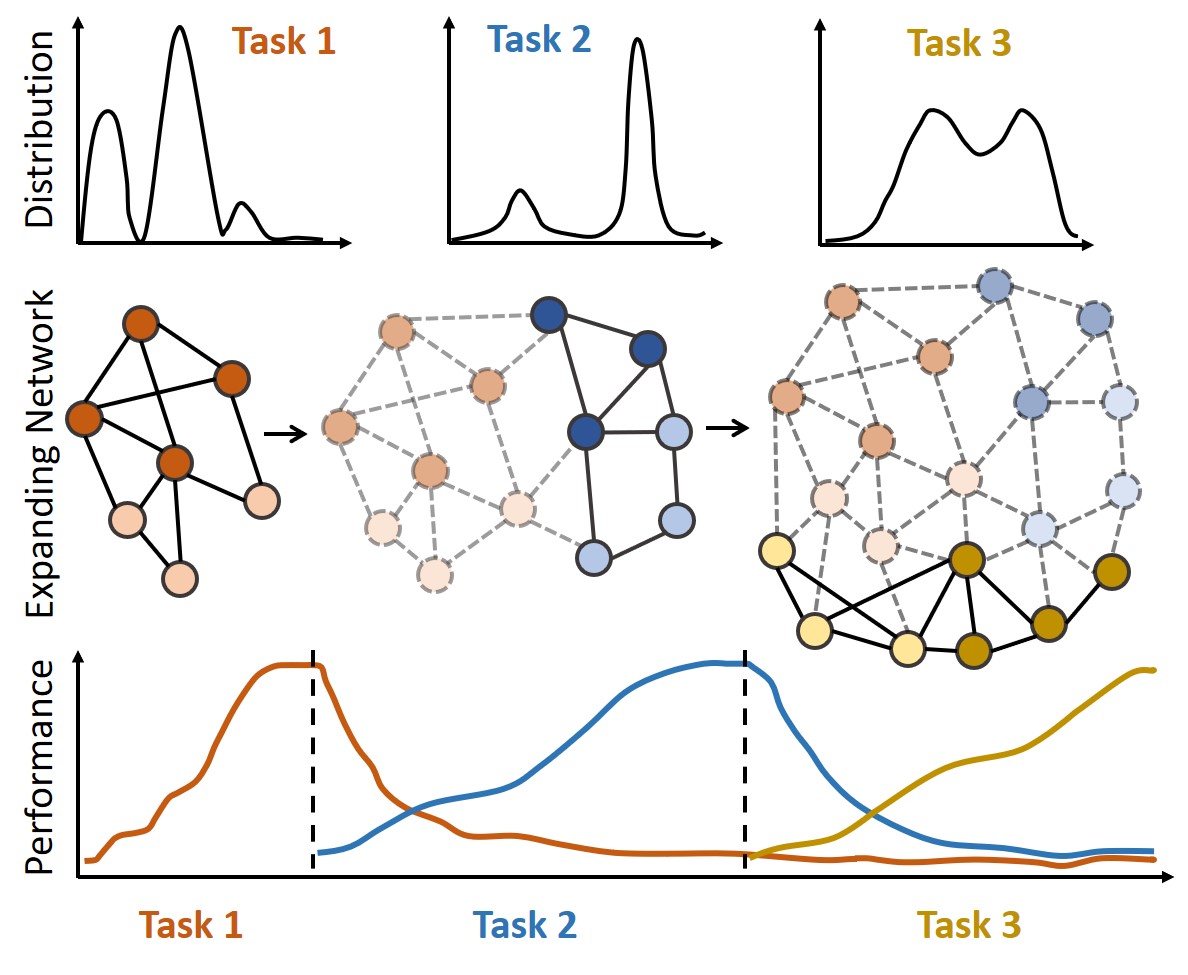} 
    \caption{Learning dynamics in an expanding network. We depict new types of nodes with different colors. The new task consisting of new types of nodes may exhibit a different distribution from existing ones. Consequently, as the model adapts to these new types of nodes, it may undergo a significant performance degradation on existing tasks, a phenomenon known as catastrophic forgetting.}
    \label{fig:intro}
\end{figure}

{Traditional machine learning techniques for networks typically assume the types of nodes and their associated edges to be static}~\footnote{Network is a type of graph. These two terms may be used interchangeably}. However, real-world networks often expand constantly with emerging new types of nodes and their associated edges. Consequently, models trained incrementally on the new node types may experience catastrophic forgetting (severe performance degradation) on the old ones as shown in Figure \ref{fig:intro}. Targeting this challenge, 
continual learning on expanding networks~\citep{liu2021overcoming,zhou2021overcoming,zhang2023hpns} has attracted increasingly more attention recently. It exhibits enormous value in various practical applications, especially in the case where networks are relatively large, and retraining a new model over the entire network is computationally infeasible.  For instance, in a social network, a community detection model has to keep adapting its parameters based on nodes from newly emerged communities; in a citation network, a document classifier needs to continuously update its parameters to distinguish the documents of newly emerged research fields. 

\begin{figure*}
    \centering
    \includegraphics[height=5.9cm]{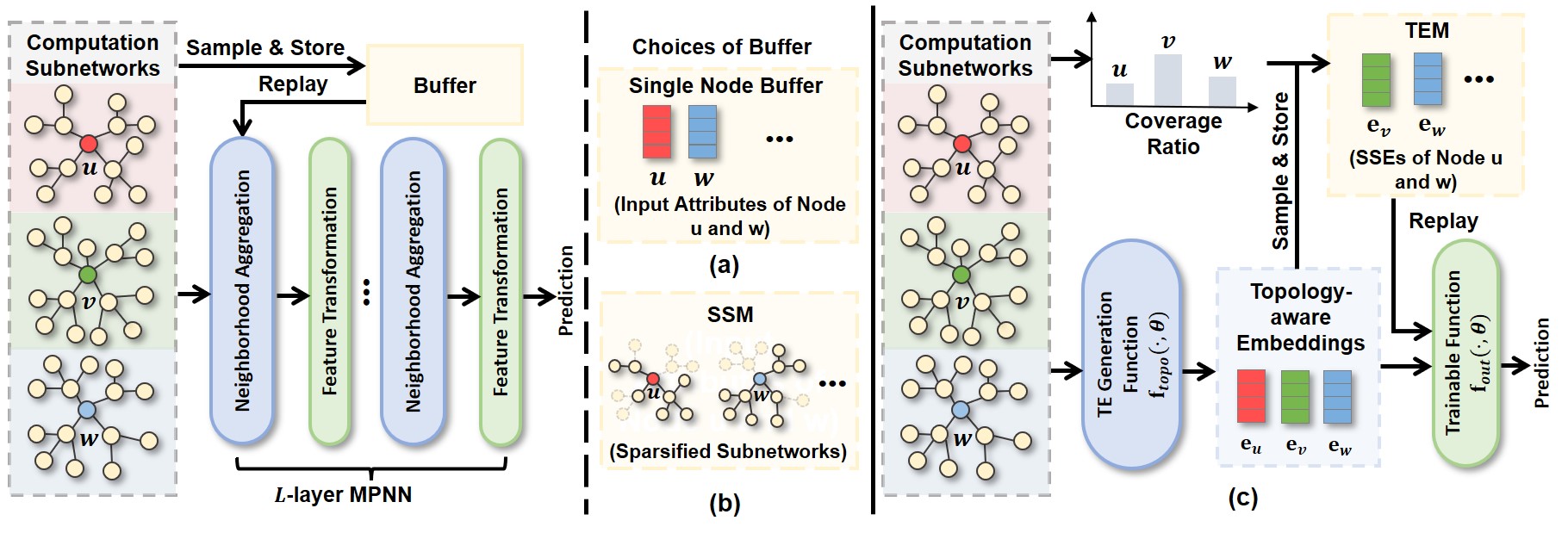}
    \caption{(a) ER-GNN \citep{zhou2021overcoming} that stores the \textit{input attributes} of individual nodes. (b) Sparsified Subgraph Memory (SSM)~\citep{zhang2022sparsified} that stores sparsified computation ego-subnetworks. (c) Our PDGNNs with TEM. The incoming computation ego-subnetworks are embedded as TEs and then fed into the trainable function. The stored TEs are sampled based on their coverage ratio (Section~\ref{sec: cover max sample}).}
    \label{fig:pipeline_3}
\end{figure*}

Memory replay~\citep{rebuffi2017icarl,lopez2017gradient,aljundi2019gradient,shin2017continual}, which stores representative examples in a buffer to retrain the model and maintain its performance over existing tasks, exhibits great success in preventing catastrophic forgetting for various continual learning tasks, \textit{e.g.}, computer vision and reinforcement learning \citep{kirkpatrick2017overcoming,li2017learning,aljundi2018memory,rusu2016progressive}. Directly applying memory replay to network data with the popular message passing neural networks (MPNNs, the general framework for most GNNs) \citep{gilmer2017neural,kipf2017semisupervised,velivckovic2017graph}, however, could give rise to the memory explosion problem because the necessity to consider the explicit topological information of target nodes. Specifically, due to the message passing over the topological connections in networks, retraining an $L$-layer GNN (Figure~\ref{fig:pipeline_3}, left) with $n$ buffered nodes would require storing $\mathcal{O}(nd^L)$ nodes \citep{chiang2019cluster,chen2017stochastic} (the number of edges is not counted yet) in the buffer, where $d$ is the average node degree.
Take the Reddit dataset~\citep{hamilton2017inductive} as an example, its average node degree is 492, and the buffer size will easily be intractable even with a 2-layer GNN. To resolve this issue, Experience Replay based GNN (ER-GNN) \citep{zhou2021overcoming}  stores representative input nodes (\textit{i.e.}, node attributes) in the buffer but ignores the topological information (Figure~\ref{fig:pipeline_3} a). Feature graph network (FGN) \citep{wang2020bridging} implicitly encodes node proximity with the inner products between the features of the target node and its neighbors. However, the explicit topological connections are abandoned and message passing is no longer feasible on the graph. Sparsified Subgraph Memory (SSM)~\citep{zhang2022sparsified} and Subgraph Episodic Memory (SEM-curvature)~\cite{zhang2023ricci} sparsify the computation ego-subnetworks for tractable memory consumption, which still partially sacrifices topological information, especially when the computation ego-subnetworks are large and a majority of nodes/edges is removed after sparsification (Figure~\ref{fig:pipeline_3} b).

To this end, \textcolor{black}{we present a general framework of Parameter Decoupled Graph Neural Networks (PDGNNs) with Topology-aware Embedding Memory (TEM) for continual learning on expanding networks (Figure~\ref{fig:pipeline_3} c).}
First, we demonstrate that the necessity to store the complete computation ego-subnetworks arises from the entanglement between the trainable parameters and the individual nodes/edges (Section~\ref{sec:Memory Replay Meets GNNs}). Targeting this problem, we design the PDGNNs, which decouple the trainable parameters from individual nodes/edges. PDGNNs enable us to develop a novel concept, \textit{Topology-aware Embedding} (TE), which is a vector with a fixed size but contains all necessary information for retraining PDGNNs. Such TEs are desired surrogates of computation ego-subnetworks to facilitate memory replay. After learning each task, a subset of TEs is sampled and stored in the \textit{Topology-aware Embedding Memory} (TEM).
Because the size of a TE is fixed, the space complexity of a memory buffer with size $n$ can be dramatically reduced from $\mathcal{O}(nd^L)$ to $\mathcal{O}(n)$. Moreover, different from continual learning on independent data without topology (\textit{e.g.}, images), we theoretically discover that replaying the TE of a single node incurs a \textit{pseudo-training effect} on its neighbors, which also alleviate the forgetting problem for the other nodes in the same computation ego-subnetwork. 
Pseudo-training effect suggests that TEs with larger coverage ratio are more beneficial to continual learning. Based on the theoretical finding, we develop the \textit{coverage maximization sampling} strategy, which effectively enhances the performance for a tight memory budget. In our experiments, 
thorough empirical studies demonstrate that PDGNNs-TEM outperform the state-of-the-art methods in both class-incremental (class-IL) \citep{rebuffi2017icarl,zhang2022sparsified,zhang2022cglb} and the task-incremental (task-IL) continual learning scenarios \citep{liu2021overcoming,zhou2021overcoming}.

\section{Related Works}

\subsection{Continual Learning \& Continual Learning on Expanding Networks}\label{sec:continual learning related works}
Existing continual learning (CL) approaches can be categorized into regularization, memory replay, and parameter isolation based methods. Regularization based methods aim to prevent drastic modification to parameters that are important for previous tasks \citep{farajtabar2020orthogonal,kirkpatrick2017overcoming,li2017learning,aljundi2018memory,hayes2020lifelong,rakaraddi2022reinforced,sun2022self,qidata,lin2023gated,sun2023self,wu2021striking,wang2018cosface,gidaris2018dynamic}. Parameter isolation methods adaptively allocate new parameters for the new tasks to protect the ones for the previous tasks \citep{wortsman2020supermasks,wu2019large,yoon2019scalable,yoon2017lifelong,rusu2016progressive,zhang2023continual}. Memory replay based methods store and replay representative data from previous tasks when learning new tasks \citep{caccia2020online,chrysakis2020online,rebuffi2017icarl,lopez2017gradient,aljundi2019gradient,shin2017continual,zhang2023ricci}.
Recently, CL on expanding networks attracts increasingly more attention due to its practical importance~ \citep{zhang2024continual,das2022graph,das2022online,das2022learning,wang2020lifelong,xu2020graphsail,daruna2021continual,carta2021catastrophic,zhang2024topology,zhang2023hpns,zhang2022cglb,febrinanto2023graph,yuan2023continual,cai2022multimodal,wei2022incregnn,wang2022streaming,ahrabian2021structure,su2023towards,kim2022dygrain,10191071,GALKE2023156,Li_Wang_Luo_Liu_Ke_2023,cui2023lifelong,liu2023cat}. Existing methods include regularization ones like topology-aware weight preserving (TWP) \citep{liu2021overcoming} that preserves crucial topologies, parameter isolation methods like HPNs \citep{zhang2023hpns} that select different parameters for different tasks, and memory replay methods like ER-GNN \citep{zhou2021overcoming}, SSM~\citep{zhang2022sparsified}, and SEM-curvature\cite{zhang2023ricci} that store representative nodes or sparsified computation ego-subnetworks. Our work is also memory based and its key advantage is the capability to preserve complete topological information with reduced space complexity, which shows significant superiority in class-IL setting (Section \ref{sec:Class-IL and Task-IL}). 
Finally, it is worth highlighting the difference between CL on expanding networks and some relevant research areas. First, dynamic graph learning \citep{9533412,wang2020streaming,han2020graph,yu2018netwalk,nguyen2018continuous,zhou2018dynamic,ma2020streaming,feng2020incremental,jiang2024empowering} focuses on the temporal dynamics with all previous data being accessible. In contrast, CL on expanding networks aims to alleviate forgetting, therefore the previous data is inaccessible. Second, few-shot graph learning \citep{zhou2019meta,guo2021few,yao2020graph,tan2022graph} targets fast adaptation to new tasks. In training, few-shot learning models can access all previous tasks (unavailable in CL). In testing, few-shot learning models need to be fine-tuned on the test classes, while the CL models are tested on existing tasks without fine-tuning. 

\subsection{GNNs $\&$ Reservoir Computing}\label{sec:decoupled GNN related works}
Graph Neural Networks (GNNs) are deep learning models designed to generate representations for graph data, which typically interleave the neighborhood aggregation and node feature transformation to extract the topological features~\citep{kipf2017semisupervised,wang2020microsoft,gilmer2017neural,velivckovic2017graph,xu2018powerful,chen2018fastgcn,hamilton2017inductive,yan2022two,yang2023simple,zhang2022graph,zhang2019graph,zhang2020context,wu2021gait,chen2009efficient,wang2016bring}. GNNs without interleaving the neighborhood aggregation and node feature transformation have been developed to reduce the computation complexity and increase the scalability~\citep{zeng2021decoupling,chen2020graph, chen2019powerful,nt2019revisiting,frasca2020sign,fey2021gnnautoscale,cong2020minimal,dong2021equivalence,zhang2020dropping}.  For example, Simple Graph Convolution (SGC) \citep{wu2019simplifying} removes the non-linear activation from GCN~\citep{kipf2017semisupervised} and only keeps one neighborhood aggregation and one node transformation layer. Approximate Personalized Propagation of Neural Predictions (APPNP)~\citep{klicpera2018predict} first performs node transformation and then conducts multiple neighborhood aggregations in one layer. 
Motivated by these works, the PDGNNs framework in this paper is specially designed to decouple the neighborhood aggregation with trainable parameters, and derive the topology-aware embeddings (TEs) to reduce the memory space complexity and facilitate continual learning on expanding networks. 
Besides, PDGNNs are also related to reservoir computing \citep{gallicchio2020fast,gallicchio2010graph}, which embed the input data (\textit{e.g.} graphs) via a fixed non-linear system. The reservoir computing modules can be adopted in PDGNNs (Equation~\ref{eq:sse_func}). 


\section{Parameter Decoupled GNNs with Topology-aware Embedding Memory}
In this section, we first introduce the notations, and then explain the technical challenge of applying memory replay techniques to GNNs. Targeting the challenge, we introduce PDGNNs with Topology-aware Embedding Memory (TEM). \textcolor{black}{Finally, inspired by theoretical findings of the \textit{pseduo-training effect}, we develop the coverage maximization sampling to enhance the performance when the memory budget is tight, which has shown its effectiveness in our empirical study.} All detailed proofs are provided in Appendix~\ref{sec:additonal theoretical results}.

\subsection{Preliminaries}\label{sec:preliminaries}
Continual learning on expanding networks is formulated as learning node representations on a sequence of subnetworks (tasks): $\mathcal{S}$ = $\{\mathcal{G}_1,\mathcal{G}_2, ..., \mathcal{G}_T\}$. Each $\mathcal{G}_{\tau}$ (\textit{i.e.}, $\tau$-th task) contains several new categories of nodes in the overall network, and is associated with a node set $\mathbb{V}_{\tau}$ and an edge set $\mathbb{E}_{\tau}$, which is represented as the adjacency matrix $\mathbf{A}_{\tau}\in \mathbb{R}^{|\mathbb{V}_{\tau}|\times|\mathbb{V}_{\tau}|}$. $\mathbb{V}$ will be used to denote an arbitrary node set in the following. 
The degree of a node $d$ refers to the number of edges connected to it. In practice, $\mathbf{A}_{\tau}$ is often normalized as $\hat{\mathbf{A}}_{\tau} = \mathbf{D}_{\tau}^{-\frac{1}{2}}\mathbf{A}_{\tau}\mathbf{D}_{\tau}^{-\frac{1}{2}}$, where $\mathbf{D}_{\tau}\in \mathbb{R}^{|\mathbb{V}_{\tau}|\times|\mathbb{V}_{\tau}|}$ is the degree matrix. Each node $v\in\mathbb{V}_{\tau}$ has a feature vector $\mathbf{x}_v \in \mathbb{R}^b$.
In classification tasks, each node $v$ has a label $\mathbf{y}_v \in \{0,1\}^C$, where $C$ is the total number of classes. 
When generating the representation for a target node $v$, a L-layer GNN typically takes a computation ego-subnetwork $\mathcal{G}^{sub}_{{\tau},v}$, containing the $L$-hop neighbors of $v$ (\textit{i.e.} $\mathcal{N}^L(v)$), as the input. For simplicity, $\mathcal{G}^{sub}_{v}$ is used in the following.

\subsection{Memory Replay Meets GNNs}\label{sec:Memory Replay Meets GNNs}
In traditional continual learning, a model $\mathrm{f}(\cdot ; \boldsymbol{\theta})$ parameterized by $\boldsymbol{\theta}$ is sequentially trained on $T$ tasks. Each task $\tau$ ($\tau \in \{1,...,T\}$) corresponds to a dataset $\mathbb{D}_{\tau} = \{(\mathbf{x}_i, \mathbf{y}_i)_{i=1}^{n_{\tau}}\}$. To avoid forgetting, memory replay based methods store representative data from the old tasks in a buffer $\mathcal{B}$. When learning new tasks. A common approach to utilize $\mathcal{B}$ is through an auxiliary loss:
\begin{align}\label{eq:aux_loss}
    \mathcal{L} = \underbrace{\sum_{\mathbf{x}_i \in \mathbb{D}_{\tau}} l(\mathrm{f}(\mathbf{x}_i; \boldsymbol{\theta}), \mathbf{y}_i)}_{\text{$\mathcal{L}_{\tau}$: loss of the current task}} 
    + \lambda \underbrace{\sum_{\mathbf{x}_j \in \mathcal{B}} l(\mathrm{f}(\mathbf{x}_j; \boldsymbol{\theta}), \mathbf{y}_j)}_{\text{$\mathcal{L}_{aux}$: auxiliary loss}},
\end{align}
where $l(\cdot, \cdot)$ denotes the loss function, and $\lambda\geq 0$ balances the contribution of the old data. Instead of directly minimizing $\mathcal{L}_{aux}$, the buffer $\mathcal{B}$ may also be used in other ways to prevent forgetting~\citep{lopez2017gradient,rebuffi2017icarl}.
In these applications, the space complexity of a buffer containing $n$ examples is $\mathcal{O}(n)$.

However, to capture the topological information, GNNs obtain the representation of a node $v$ based on a computation ego-subnetwork surrounding $v$. \textcolor{black}{We exemplify it with the popular MPNN framework \citep{gilmer2017neural}}, which updates the hidden node representations at the $l+1$-th layer as:
\begin{align}
    &\mathbf{m}_v^{l+1} = \sum_{w\in \mathcal{N}^1(v)} \mathrm{M}_l (\mathbf{h}^l_v, \mathbf{h}^l_w, \mathbf{x}^e_{v,w}; \boldsymbol{\theta}_l^\mathrm{M}),&
    &\mathbf{h}^{l+1}_v = \mathrm{U}_l(\mathbf{h}_v^l,\mathbf{m}_v^{l+1}; \boldsymbol{\theta}_l^\mathrm{U}),&
\end{align}
where $\mathbf{h}_v^l$, $\mathbf{h}_w^l$ are hidden representations of nodes at layer $l$, $\mathbf{x}^e_{v,w}$ is the edge feature, $\mathrm{M}_l (\cdot, \cdot, \cdot; \boldsymbol{\theta}_l^\mathrm{M})$ is the message function to integrate neighborhood information, and $\mathrm{U}_l(\cdot,\cdot; \boldsymbol{\theta}_l^\mathrm{U})$ updates $\mathbf{m}_v^{l+1}$ into $\mathbf{h}_v^l$ ($\mathbf{h}_v^0$ is the input features).
In a $L$-layer MPNN, the representation of a node $v$ can be simplified as,
\begin{align}
    \mathbf{h}^{L}_v = \mathrm{MPNN}(\mathbf{x}_v, \mathcal{G}^{sub}_v; \boldsymbol{\Theta}),
\end{align}
where $\mathcal{G}^{sub}_v$ 
contains the $L$-hop neighbors ($\mathcal{N}^L(v)$), $\mathrm{MPNN}(\cdot, \cdot; \boldsymbol{\Theta})$ is the composition of all
$\mathrm{M}_l (\cdot, \cdot, \cdot; \boldsymbol{\theta}_l^\mathrm{M})$ and $\mathrm{U}_l(\cdot,\cdot; \boldsymbol{\theta}_l^\mathrm{U})$ at different layers. Since $\mathcal{N}^L(v)$ typically contains $\mathcal{O}(d^L)$ nodes, replaying $n$ nodes requires storing $\mathcal{O}(nd^L)$ nodes (the edges are not counted yet), where $d$ is the average degree. 
\color{black}Therefore, the buffer size will be easily intractable in practice (\textit{e.g.} the example of Reddit dataset in Introduction), and directly storing the computation ego-subnetworks
for memory replay is infeasible for GNNs.
\color{black}
\subsection{Parameter Decoupled GNNs with TEM }\label{sec:pdgnn with ssem}

As we discussed earlier, the key challenge of applying memory replay to network data is to preserve the rich topological information of the computation ego-subnetworks with potentially unbounded sizes. Therefore, a natural resolution is to preserve the crucial topological information with a compact vector such that the memory consumption is tractable. Formally, the desired subnetwork representation can be defined as \textit{Topology-aware Embedding} (TE).
\color{black}
\begin{definition}[Topology-aware embedding]\label{def:sse}
Given a specific GNN parameterized with $\boldsymbol{\theta}$ and an input $\mathcal{G}^{sub}_v$, an embedding vector $\mathbf{e}_v$ is a topology-aware embedding for $\mathcal{G}^{sub}_v$ with respect to this GNN, if optimizing $\boldsymbol{\theta}$ with $\mathcal{G}^{sub}_v$ or $\mathbf{e}_v$ for this specific GNN are equivalent, i.e. $\mathbf{e}_v$ contains all necessary topological information of $\mathcal{G}^{sub}_v$ for training this GNN. 
\end{definition}
\color{black}
However, TEs cannot be directly derived from the MPNNs due to their interleaved neighborhood aggregation and feature transformations. According to Section~\ref{sec:Memory Replay Meets GNNs}, whenever the trainable parameters get updated, recalculating the representation of a node $v$ requires all nodes and edges in $\mathcal{G}^{sub}_v$. To resolve this issue, we formulate the Parameter Decoupled Graph Neural Networks (PDGNNs) framework, which decouples the trainable parameters from the individual nodes/edges. PDGNNs may not be the only feasible framework to derive TEs, but is the first attempt and is empirically effective. Given $\mathcal{G}^{sub}_v$, the prediction of node $v$ with PDGNNs consists of two steps. First, the topological information of $\mathcal{G}^{sub}_v$ is encoded into an embedding $\mathbf{e}_v$ via the function $\mathrm{f}_{topo} (\cdot)$ without trainable parameters (instantiations of $\mathrm{f}_{topo} (\cdot)$ are detailed in Section~\ref{sec:Instantiations of PDGNNs}).
\begin{align}\label{eq:sse_func}
    &\mathbf{e}_v =\mathrm{f}_{topo} (\mathcal{G}^{sub}_v).
\end{align}
Next, $\mathbf{e}_v$ is further passed into a trainable function $\mathrm{f}_{out}(\cdot; \boldsymbol{\theta})$ parameterized by $\boldsymbol{\theta}$ (instantiations of $\mathrm{f}_{out}(\cdot; \boldsymbol{\theta})$ are detailed in Section~\ref{sec:Instantiations of PDGNNs}) to get the output prediction $\mathbf{\hat{y}}_v$,
\begin{align}
    &\mathbf{\hat{y}}_v = \mathrm{f}_{out}(\mathbf{e}_v; \boldsymbol{\theta}).
\end{align}
With the formulations above, $\mathbf{e}_v$ derived in Eq. (\ref{eq:sse_func}) clearly satisfies the requirements of TE (Definition~\ref{def:sse}). Specifically, since the trainable parameters acts on $\mathbf{e}_v$ instead of any individual node/edge, optimizing the model parameters $\boldsymbol{\theta}$ with either $\mathbf{e}_v$ or $\mathcal{G}^{sub}_v$ are equivalent. 
Therefore, to retrain the model, the memory buffer only needs to store TEs instead of the original computation ego-subnetworks, which reduces the space complexity from $\mathcal{O}(nd^L)$ to $\mathcal{O}(n)$. 
We name the buffer to store the TEs as \textit{Topology-aware Embedding Memory} ($\mathcal{TEM}$). Given a new task $\tau$, the update of $\mathcal{TEM}$ is: 
\begin{align}\label{eq:ssem}
    \mathcal{TEM}=\mathcal{TEM} \; \bigcup \; \mathrm{sampler}(\{\mathbf{e}_v \mid v\in \mathbb{V}_{\tau}\}, n),
\end{align}
where $\mathrm{sampler}(\cdot,\cdot)$ is the adopted sampling strategy to populate the buffer, $\bigcup$ denotes the set union, and $n$ is the budget. \textcolor{black}{According to the experimental results (Section~\ref{sec:Performance vs. Coverage Ratio and Studies on the Buffer Size}), as long as $\mathcal{TEM}$ is maintained, PDGNNs-TEM can perform reasonably well with different choices of $\mathrm{sampler}(\cdot,\cdot)$, including the random sampling. Nevertheless, in Section \ref{sec: cover max sample}, based on the theoretical insights in Section \ref{sec:bene_side_eff_TEs}, we propose a novel sampling strategy to better populate $\mathcal{TEM}$ when the memory budget is tight, which is empirically verified to be effective in Section~\ref{sec:Performance vs. Coverage Ratio and Studies on the Buffer Size}.}
Besides, Equation~(\ref{eq:ssem}) assumes that all data of the current task are presented concurrently. In practice, the data of a task may come in multiple batches (\textit{e.g.}, nodes come in batches on large networks), and the buffer update have to be slightly modified by either storing sizes of the computational ego-networks and recalculating the multinomial distribution or adopting reservoir sampling. 
For task $\tau$ with network $\mathcal{G}_{\tau}$, the loss with $\mathcal{TEM}$ then becomes:
\begin{align}\label{eq:new loss}
    \mathcal{L} = \underbrace{\sum_{v \in \mathbb{V}_{\tau}} l(\mathrm{f}_{out}(\mathbf{e}_v; \boldsymbol{\theta}), \mathbf{y}_v)}_{\text{$\mathcal{L}_{\tau}$: loss of the current task $\tau$}} + \underbrace{\lambda\sum_{\mathbf{e}_w \in \mathcal{TEM}} l(\mathrm{f}_{out}(\mathbf{e}_w;\boldsymbol{\theta}), \mathbf{y}_w)}_{\text{$\mathcal{L}_{aux}$: auxiliary loss}}. 
\end{align}
$\lambda$ balances the contribution of the data from the current task and the memory, and is typically manually chosen in traditional continual learning works. However, on network data, we adopt a different strategy to re-scale the losses according to the class sizes to counter the bias from the severe class imbalance, which cannot be handled on networks by directly balancing the datasets. 


\subsection{Instantiations of PDGNNs}\label{sec:Instantiations of PDGNNs}
Although without trainable parameters, the function $\mathrm{f}_{topo} (\cdot)$ for generating TEs can be highly expressive with various formulations including linear and non-linear ones, both of which are studied in this work. 
First, the linear instantiations of $\mathrm{f}_{topo} (\cdot)$ can be generally formulated as, 
\color{black}
\begin{align}\label{eq:sse}
    \mathbf{e}_v=\mathrm{f}_{topo} (\mathcal{G}^{sub}_v) = \sum_{w \in \mathbb{V}} \mathbf{x}_w \cdot \pi(v, w; \hat{\mathbf{A}}), 
\end{align}
where $\pi(\cdot, \cdot; \hat{\mathbf{A}})$ denotes the strategy for computation ego-subnetwork construction and determines how would the model capture the topological information. Equation~(\ref{eq:sse}) describes the operation on each node. In practice, Equation~(\ref{eq:sse}) could be implemented as matrix multiplication to generate TEs of a set of nodes $\mathbb{V}$ in parallel, \textit{i.e.} $\mathbf{E}_{\mathbb{V}} = \boldsymbol{\Pi}  \mathbf{X}_{\mathbb{V}}$, where each entry $\boldsymbol{\Pi}_{v,w}=\pi(v, w; \hat{\mathbf{A}})$. $\mathbf{E}_{\mathbb{V}}\in \mathbb{R}^{|\mathbb{V}|\times b}$ is the concatenation of all TEs ($\mathbf{e}_v \in \mathbb{R}^b$), and $\mathbf{X}_{\mathbb{V}} \in \mathbb{R}^{|\mathbb{V}|\times b}$ is the concatenation of all node feature vectors $\mathbf{x}_v \in \mathbb{R}^b$. In our experiments, we adopt three representative strategies. The first strategy (S1)~\citep{wu2019simplifying} is a basic version of message passing and can be formulated as $\boldsymbol\Pi = \hat{\mathbf{A}}^L$. The second strategy (S2)~\citep{zhu2020simple} considers balancing the contribution of neighborhood information from different hops via a hyperparameter $\alpha$, \textit{i.e.} $\boldsymbol\Pi=\frac{1}{L}\sum_{l=1}^{L}\big( (1-\alpha)\hat{\mathbf{A}}^l + \alpha \mathbf{I} \big)$. Finally, we also adopt a strategy (S3)~\citep{klicpera2018predict} that adjusts the contribution of the neighbors based on PageRank~\citep{page1999pagerank}, \textit{i.e.} $\boldsymbol\Pi=\big((1-\alpha)\hat{\mathbf{A}}+\alpha \mathbf{I}\big)^L$, in which $\alpha$ also balances the contribution of the neighborhood information.

The linear formulation of $\mathrm{f}_{topo} (\cdot)$ (Equation~(\ref{eq:sse})) yields both promising experimental results (Section \ref{sec:experiments}) and instructive theoretical results (Section \ref{sec:bene_side_eff_TEs}, and \ref{sec: cover max sample}). Equation~(\ref{eq:sse}) is also highly efficient especially for large networks due to the absence of iterative neighborhood aggregations. But $\mathrm{f}_{topo} (\cdot)$ can also take non-linear forms with more complex mappings. 
\color{black}
For example, we can also adopt a reservoir computing module \citep{gallicchio2020fast} to instantiate $\mathrm{f}_{topo} (\cdot)$, which is formulated as,
\begin{align}
    \mathrm{f}_{topo}^i (\mathcal{G}^{sub}_v) = \mathrm{tanh}\Big(\mathbf{W}_I \cdot \mathbf{x}_v^{i-1} +  \sum_{w \in \mathbb{V}} \mathbf{W}_H \cdot \mathbf{x}_w^{i-1} \Big) , i=1,..,L, 
\end{align}
where $\mathbf{W}_I$ and $\mathbf{W}_H$ are fixed weight matrices, and $\mathbf{e}_v = \mathrm{f}_{topo}^L(\mathcal{G}^{sub}_v)$. 
\color{black}

Since $\mathrm{f}_{out}(\cdot; \boldsymbol{\theta})$ simply deals with individual vectors (TEs), it is instantiated as MLP in this work. The specific configurations of $\mathrm{f}_{out}(\cdot; \boldsymbol{\theta})$ is described in the experimental part (Section~\ref{sec:experiment setup}).
\color{black}


\begin{figure}
  \begin{center}
    \includegraphics[width=0.48\textwidth]{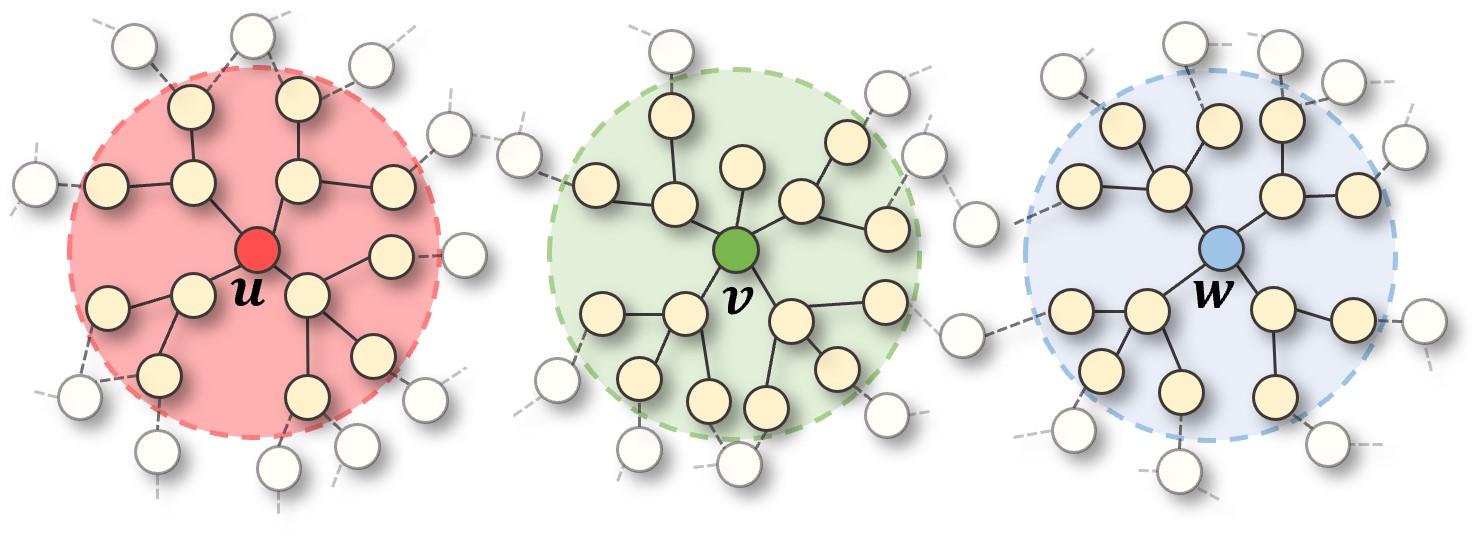}
  \end{center}
  \caption{Illustration of the coverage ratio. Supposing the network has $N$ nodes, $R_c(\{u\})=\frac{13}{N}$,  $R_c(\{v\})=\frac{15}{N}$,  $R_c(\{u\})=\frac{14}{N}$, and  $R_c(\{u,v,w\})=\frac{42}{N}$}
  \label{fig:cover_ratio}
\end{figure}

\subsection{Pseudo-training Effects of TEs}\label{sec:bene_side_eff_TEs}
In traditional continual learning \textcolor{black}{on independent data without explicit topology}, replaying an example $\mathbf{x}_i$ \textcolor{black}{(\textit{e.g.}, an image)} only reinforces the prediction of itself. In this subsection, we introduce the pseudo-training effect, which implies that training PDGNNs with $\mathbf{e}_v$ of node $v$ also influences the predictions of the other nodes in $\mathcal{G}^{sub}_v$, based on which we develop a novel sampling strategy to further boost the performance with a tight memory budget.

\begin{thm}[Pseudo-training]\label{thm:equivalence}
Given a node $v$, its computation ego-subnetwork $\mathcal{G}^{sub}_v$, the TE $\mathbf{e}_v$, and label $\mathbf{y}_v$ (suppose $v$ belongs to class $k$, \textit{i.e.} $\mathbf{y}_{v,k}=1$), then training PDGNNs with $\mathbf{e}_v$ has the following two properties:
  
$\mathrm{1}$. It is equivalent to training PDGNNs with each node $w$ in $\mathcal{G}^{sub}_v$ with $\mathcal{G}^{sub}_v$ being a pseudo computation ego-subnetwork and $\mathbf{y}_v$ being a pseudo label, where the contribution of $\mathbf{x}_w$ (via Equation~\ref{eq:sse}) is re-scaled by $\frac{\pi(v, w; \hat{\mathbf{A}})}{\pi(w, w; \hat{\mathbf{A}})}$. We term this property as the pseudo-training effect on neighboring nodes, \textcolor{black}{because it is equivalent to that the training is conducted on each neighboring node (in $\mathcal{G}^{sub}_v$) through the pseudo labels and the pseudo computation ego-subnetworks.}

$\mathrm{2}$. When $\mathrm{f}_{out}(\cdot; \boldsymbol{\theta})$ is linear, training PDGNNs on $\mathbf{e}_v$ is also equivalent to training $\mathrm{f}_{out}(\cdot; \boldsymbol{\theta})$ on pseudo-labeled nodes ($\mathbf{x}_w$, $\mathbf{y}_v$) for each $w$ in $\mathcal{G}^{sub}_v$, where the contribution of $w$ in the loss is adaptively re-scaled with a weight $\frac{\mathrm{f}_{out}(\mathbf{x}_w; \boldsymbol{\theta})_k\cdot\pi(v, w; \hat{\mathbf{A}})}{ \sum_{w \in \mathbb{V}_v^{sub}} \mathrm{f}_{out}\big(\mathbf{x}_w \cdot \pi(v, w; \hat{\mathbf{A}}); \boldsymbol{\theta}\big)_k}$.
\end{thm}

The pseudo-training effect essentially arises from the neighborhood aggregation operation of GNNs, of which the rationale is to iteratively refine the node embeddings with similar neighbors. Pseudo-training effect implies that replaying the TE of one node can also strengthen the prediction for its neighbors within the same computation ego-subnetwork and alleviate the forgetting problem on them. 
The above analysis suggests that TEs with larger computation ego-subnetworks covering more nodes may be more effective, motivating our coverage maximization sampling strategy in the next subsection, which is also empirically justified in Section~\ref{sec:Performance vs. Coverage Ratio and Studies on the Buffer Size}.

\begin{algorithm}[]
  \caption{Coverage maximization sampling}
  \label{alg:max sample}
  \begin{algorithmic}[1]
  \STATE \textbf{Input:} $\mathcal{G}_{\tau}$, $\mathbb{V}_{\tau}$, $\hat{\mathbf{A}}_{\tau}$, $\pi(\cdot,\cdot;\cdot)$, sample size $n$.
  \STATE \textbf{Output:} Selected nodes $\mathcal{S}$
  \STATE Initialize $\mathcal{S} = \{\}$.
    \FOR{each $v \in \mathbb{V}_{\tau}$}%
      \STATE $R_c( \{ v \})=\frac{|\{w|w \in \mathcal{G}_{\tau,v}^{sub}\}|}{|\mathbb{V}_{\tau}|}$
    \ENDFOR
    \FOR{each $v \in \mathbb{V}_{\tau}$}
      \STATE $p_v$ = $\frac{R_c( \{ v \})}{\sum_{w \in \mathbb{V}_{\tau}}R_c( \{ w \})}$ 
    \ENDFOR
    \WHILE{$n>0$}%
      \STATE Sample one node $v$ from $\mathbb{V}_{\tau}$ according to $\{p_w \mid w \in \mathbb{V}_{\tau}\}$.
      \STATE $\mathcal{S} = \mathcal{S} \cup \{v\}$
      \STATE $\mathbb{V}_{\tau}=\mathbb{V}_{\tau} \backslash \{v\}$ \hfill \#{Sampling without replacement}
      \STATE $n \gets n-1$ 
    \ENDWHILE
  \end{algorithmic}%
\end{algorithm}

\subsection{Pseudo-training Effect and Network Homophily}\label{proof:pseudo train and homophily}
In this subsection, we provide a brief discussion on pseudo-training and network homophily. Given a network $\mathcal{G}$, the homophily ratio is defined as the ratio of the number of edges connecting nodes with a same label and the total number of edges, \textit{i.e.} 
\begin{align}
    h(\mathcal{G}) = \frac{1}{|\mathcal{E}|}\sum_{(j,k)\in\mathcal{E}}\mathbf{1}(\mathbf{y}_j=\mathbf{y}_k),
\end{align}
where $\mathcal{E}$ is the edge set, $\mathbf{y}_j$ is the label of node $j$, and $\mathbf{1}(\cdot)$ is the indicator function \citep{ma2021homophily}. For any network, the homophily ratio is between 0 and 1. For each computation ego-subnetwork, when the homophily ratio is high, the neighboring nodes tend to share labels with the center node, and the pseudo training would be beneficial for the performance. Many real-world networks, \textit{i.e.} the social network and citation networks, tend to have high homophily ratios, and pseudo training will bring much benefit,. 

In our work, the homophily ratio of the 4 network datasets are: CoraFull-CL (0.567), Arxiv-CL (0.655), OGB-Products (0.807), Reddit-CL (0.755). These datasets cover the ones with high homophily (OGB-Products and Reddit), as well as the ones with lower homophily.

When learning on more heterophilous networks (homophily ratio close to 0) $f_{topo}(\cdot)$ is required to be constructed specially constructed.
Heterophilous network learning is largely different from homophilous network learning, and requires different GNN designs~\citep{zheng2022graph,abu2019mixhop,zhu2020beyond}. Therefore, $f_{topo}(\cdot)$ should also be instantiated to be suitable for heterophilous networks. 
The key difference of heterophilous network learning is that the nodes belonging to the same classes are not likely to be connected, and GNNs should be designed to separately process the proximal neighbors with similar information and distal neighbors with irrelevant information, or only aggregate information from the proximal neighbors~\citep{zheng2022graph,abu2019mixhop,zhu2020beyond}. 
For example, MixHop~\citep{abu2019mixhop} encodes neighbors from different hops separately. A given computation ego-subnetwork will be divided into different hops. For each hop, the model generates a separate embedding. Finally, the embeddings of different hops are concatenated as the final TE. 
H2GCN~\citep{zhu2020beyond} only aggregates higher-order neighbors that are proximal to the center node. 

In other words, via constructing $f_{topo}(\cdot)$ to be suitable for heterophilous networks, the neighborhood aggregation is still conducted on the proximal nodes, and so is the pseudo-training. In this way, the pseudo-training will still benefit the performance.

\begin{table}[]
\small
    \caption{The detailed statistics of datasets and task splittings}
    \centering
    \begin{tabular}{lcccc}\toprule
        \textbf{Dataset} & CoraFull~\cite{mccallum2000automating} &Arxiv~\cite{hu2020ogb}& Reddit~\cite{hamilton2017inductive} & Products~\cite{hu2020ogb}\\\midrule
        \# nodes         & 19,793   & 169,343  &232,965  & 2,449,029 \\\midrule
        \# edges         & 130,622  &1,166,243 & 114,615,892 &61,859,140 \\ \midrule
        \# classes       & 70  &40 & 40 & 47 \\\midrule
        \# tasks         & 30 & 20 & 20 & 23 \\\bottomrule
    \end{tabular}
    \label{tab:datasets}
\end{table}

\subsection{Coverage Maximization Sampling}\label{sec: cover max sample}

Following the above subsection, TEs with larger computation ego-subnetworks are preferred to be stored. To quantify the size of the computation ego-subnetworks, we formally define the coverage ratio of the selected TEs as the nodes covered by their computation ego-subnetworks versus the total nodes in the network (Figure~\ref{fig:cover_ratio}). Since a TE uniquely corresponds to a node, we may use `node' and `TE' interchangeably. 
\begin{definition}
Given a network $\mathcal{G}$, node set $\mathbb{V}$, and function $\pi(\cdot,\cdot;\hat{\mathbf{A}})$, the coverage ratio of a set of nodes $\mathbb{V}_s$ is:
\begin{align}\label{eq:cover_ratio}
    R_c(\mathbb{V}_s) = \frac{|\cup_{v \in \mathbb{V}_s}\{w|w \in \mathcal{G}_{v}^{sub}\}|}{|\mathbb{V}|}, 
\end{align}
\textit{i.e.}, the ratio of nodes of the entire (training) network covered by the computation ego-subnetworks of the selected nodes (TEs).
\end{definition}

\begin{table*}[]
    \caption{Performance $\&$ coverage ratios of different sampling strategies and buffer sizes on OGB-Arxiv ($\uparrow$ higher means better).}  
    \centering
    \begin{tabular}{c|c|ccccc}\toprule
     \multicolumn{2}{c|}{ Ratio of dataset /\%} &0.02&0.1&1.0&5.0&40.0\\ \midrule\midrule
        \multirow{3}{2.5em}{AA/\%}  &Uniform samp. &12.0$\pm$1.1&24.1$\pm$1.7&42.2$\pm$0.3&50.4$\pm$0.4&53.3$\pm$0.4 \\ 
         &Mean of feat. & 12.6$\pm$0.1&25.3$\pm$0.3&42.8$\pm$0.3&50.4$\pm$0.7&53.3$\pm$0.2 \\
         &Cov. Max. &\textbf{14.9$\pm$0.8}&\textbf{26.8$\pm$1.8}&\textbf{43.7$\pm$0.5}&\textbf{50.5$\pm$0.4}&\textbf{53.4$\pm$0.1} \\ \midrule
         \multirow{3}{2.5em}{Cov. \\ratio/\%} &Uniform samp. &0.1$\pm$0.1&0.3$\pm$0.0&3.5$\pm$0.9&15.9$\pm$1.1&84.8$\pm$1.5 \\ 
         &Mean of feat. &0.2$\pm$0.4&0.6$\pm$0.3&7.1$\pm$0.6&29.6$\pm$1.7&91.1$\pm$0.1 \\
         &Cov. Max. &\textbf{0.5$\pm$1.1}&\textbf{2.9$\pm$1.8}&\textbf{22.5$\pm$1.6}&\textbf{46.3$\pm$0.6}&\textbf{92.8$\pm$0.0} \\\bottomrule
    \end{tabular}
    \label{tab:performance and coverage ratio}
    \centering
    \caption{Performance comparisons under class-IL on different datasets ($\uparrow$ higher means better).}
    \begin{tabular}{c||cc|cc|cc|cc}
    \toprule
    \multirow{2}{2.5em}{\textbf{C.L.T.}}  &  \multicolumn{2}{c|}{CoraFull} &  \multicolumn{2}{c|}{OGB-Arxiv} &  \multicolumn{2}{c|}{Reddit} & \multicolumn{2}{c}{OGB-Products}\\ \cline{2-9}
                 & AA/\% $\uparrow$ & AF/\% $\uparrow$& AA/\%     $\uparrow$  & AF /\% $\uparrow$      & AA/\%  $\uparrow$     & AF /\%  $\uparrow$     & AA/\%  $\uparrow$     & AF /\% $\uparrow$\\ \bottomrule\toprule
       Fine-tune &2.9$\pm$0.0&-94.7$\pm$0.1  &4.9$\pm$0.0&-87.0$\pm$1.5  &5.1$\pm$0.3&-94.5$\pm$2.5  &3.4$\pm$0.8&-82.5$\pm$0.8\\
       EWC~(\citeyear{kirkpatrick2017overcoming}) &15.2$\pm$0.7&-81.1$\pm$1.0 &4.9$\pm$0.0&-88.9$\pm$0.3  &10.6$\pm$1.5&-92.9$\pm$1.6  &3.3$\pm$1.2&-89.6$\pm$2.0 \\
       MAS~(\citeyear{aljundi2018memory}) &12.3$\pm$3.8&-83.7$\pm$4.1 &4.9$\pm$0.0&-86.8$\pm$0.6  &13.1$\pm$2.6&-35.2$\pm$3.5  &15.0$\pm$2.1&-66.3$\pm$1.5 \\
       GEM~(\citeyear{lopez2017gradient}) &7.9$\pm$2.7&-84.8$\pm$2.7  &4.8$\pm$0.5&-87.8$\pm$0.2  &28.4$\pm$3.5&-71.9$\pm$4.2  &5.5$\pm$0.7&-84.3$\pm$0.9\\
       TWP~(\citeyear{liu2021overcoming}) &20.9$\pm$3.8&-73.3$\pm$4.1  &4.9$\pm$0.0&-89.0$\pm$0.4  &13.5$\pm$2.6&-89.7$\pm$2.7  &3.0$\pm$0.7&-89.7$\pm$1.0\\
       LwF~(\citeyear{li2017learning}) &2.0$\pm$0.2&-95.0$\pm$0.2  &4.9$\pm$0.0&-87.9$\pm$1.0 &4.5$\pm$0.5&-82.1$\pm$1.0  & 3.1$\pm$0.8&-85.9$\pm$1.4\\
       ER-GNN~(\citeyear{zhou2021overcoming}) &3.0$\pm$0.1&-93.8$\pm$0.5  &30.3$\pm$1.5&-54.0$\pm$1.3  &88.5$\pm$2.3&-10.8$\pm$2.4 &24.5$\pm$1.9&-67.4$\pm$1.9\\
       SSM~(\citeyear{zhang2022sparsified}) &{75.4$\pm$0.1}&-9.7$\pm$0.0 & {48.3$\pm$0.5}&-10.7$\pm$0.3 & {94.4$\pm$0.0}&-1.3$\pm$0.0 & {63.3$\pm$0.1}&-9.6$\pm$0.3\\
       SEM-curvature~(\citeyear{zhang2023ricci})& \underline{77.7$\pm$0.8}&-10.0$\pm$1.2& \underline{49.9$\pm$0.6}&-8.4$\pm$1.3 &\textbf{96.3$\pm$0.1}&-0.6$\pm$0.1 &\underline{65.1$\pm$1.0}&{-9.5$\pm$0.8 }\\
       \midrule
       Joint &80.6$\pm$0.3&-  &46.4$\pm$1.4&-  &99.3$\pm$0.2&-  &71.5$\pm$0.7&-\\  \midrule 
       \textbf{PDGNNs} &\textbf{81.9$\pm$0.1}&{-3.9$\pm$0.1} &\textbf{53.2$\pm$0.2}&{-14.7$\pm$0.2}& \underline{94.7$\pm$0.4}&{-3.0$\pm$0.4} &\textbf{73.9$\pm$0.1}&{-10.9$\pm$0.2}\\ 
    \bottomrule
    \end{tabular}
    \label{tab:comparisons with baselines class-IL}
\end{table*}

To maximize $R_c(\mathcal{TEM})$, a naive approach is to first select the TE with the largest coverage ratio, and then iteratively incorporate TE that increases $R_c(\mathcal{TEM})$ the most. However, this requires computing $R_c(\mathcal{TEM})$ for all candidate TEs at each iteration, which is time consuming especially on large networks. Besides, certain randomness is also desired for the diversity of $\mathcal{TEM}$. 
Therefore, we propose to sample TEs based on their coverage ratio. Specifically, in task $\tau$, the probability of sampling node $v\in \mathbb{V}_{\tau}$ is $p_v = \frac{R_c( \{ v \})}{\sum_{w \in \mathbb{V}_{\tau}}R_c( \{ w \})}$. Then the nodes in $\mathbb{V}_{\tau}$ are sampled according to $\{p_v \mid v\in \mathbb{V}_{\tau}\}$ without replacement, as shown in Algorithm \ref{alg:max sample}.
In experiments, we demonstrate the correlation between the coverage ratio and the performance, which verifies the benefits revealed in Section \ref{sec:bene_side_eff_TEs}

\section{Experiments}\label{sec:experiments}

In this section, we aim to answer the following research questions: RQ1: Whether PDGNNs-TEM works well with a reasonable buffer size? RQ2: Does coverage maximization sampling ensure a higher coverage ratio and better performance when the memory budget is tight? RQ3: Whether our theoretical results can be reflected in experiments? RQ4: Whether PDGNNs-TEM can outperform the state-of-the-art methods in both class-IL and task-IL scenarios? RQ5: How to interpret the learned node embedding under continual learning setting.
Due to the space limitations, only the most prominent results are presented in the main content.
For simplicity, PDGNNs-TEM will be denoted as PDGNNs in this section. All codes are available at \url{github.com/imZHANGxikun/PDGNNs}.

\subsection{Datasets}
Following the public benchmark CGLB~\citep{zhang2022cglb}, we adopted four datasets, CoraFull \cite{mccallum2000automating}, OGB-Arxiv \cite{hu2020ogb}, Reddit \cite{hamilton2017inductive}, and OGB-Products \cite{hu2020ogb}, with up to millions of nodes and 70 classes. 
Dataset statistics and task splittings are summarized in Table \ref{tab:datasets}. 




\subsection{Experimental Setup and Model Evaluation}\label{sec:experiment setup}

\textbf{Continual learning setting and model evaluation.}\label{sec:continual learning setting}
During training, a model is trained on a task sequence. During testing, the model is tested on all learned tasks. Class-IL requires a model to classify a given node by picking a class from all learned classes (more challenging), while task-IL only requires the model to distinguish the classes within each task. For model evaluation, the most thorough metric is the accuracy matrix $\mathrm{M}^{acc}\in \mathbb{R}^{T \times T}$, where $\mathrm{M}^{acc}_{i,j}$ denotes the accuracy on task $j$ after learning task $i$. 
\color{black}
The learning dynamics can be reflected with average accuracy (AA) over all learnt tasks after learning each task, \textit{i.e.}, $\Big\{\frac{\sum_{j=1}^i \mathrm{M}^{acc}_{i,j}}{i} | i=1,...,T \Big\}$,
which can be visualized as a curve.
Similarly, the average forgetting (AF) after learning each task reflects the learning dynamics from the perspective of forgetting, $\Big\{\frac{\sum_{j=1}^{i-1} \mathrm{M}^{acc}_{i,j}-\mathrm{M}^{acc}_{j,j}}{i-1} | i=2,...,T \Big\}$.
\color{black}
To use a single numeric value for evaluation, the AA and AF after learning all $T$ tasks will be used. These metrics are widely adopted in continual learning works~\citep{chaudhry2018riemannian,lopez2017gradient,liu2021overcoming,zhang2023hpns,zhou2021overcoming}, although the names are different in different works. 
We repeat all experiments 5 times on one Nvidia Titan Xp GPU. All results are reported with average performance and standard deviations.

\noindent\textbf{Baselines and model settings.}\label{sec:Baselines and model settings}
Our baselines for continual learning on expanding networks include Experience Replay based GNN (ER-GNN) \citep{zhou2021overcoming}, Topology-aware Weight Preserving~(TWP) \citep{liu2021overcoming}, Sparsified Subgraph Memory (SSM)~\citep{zhang2022sparsified}, and Subgraph Episodic Memory (SEM)~\cite{zhang2023ricci}. Milestone works for Euclidean data but also applicable to GNNs include Elastic Weight Consolidation~(EWC) \citep{kirkpatrick2017overcoming}, Learning without Forgetting~(LwF) \citep{li2017learning}, Gradient Episodic Memory~(GEM) \citep{lopez2017gradient}, and Memory Aware Synapses (MAS) \citep{aljundi2018memory}), are also adopted. HPNs~\citep{zhang2023hpns} is designed to work under a stricter task-IL setting, and cannot be properly incorporated for comparison. The results of the baselines are adopted from the original works \cite{zhang2022cglb,zhang2022sparsified,zhang2023ricci}. 
Besides, joint training (without forgetting problem) and fine-tune (without continual learning technique) are adopted as the upper and lower bound on the performance. We instantiate $\mathrm{f}_{out}(\cdot;\boldsymbol\theta)$ as a multi-layer perceptron (MLP). 
All methods including $\mathrm{f}_{out}(\cdot;\boldsymbol\theta)$ of PDGNNs are set as 2-layer with 256 hidden dimensions, and $L$ in Section~\ref{sec:pdgnn with ssem} is set as 2 for consistency. 
As detailed in Section~\ref{sec:Performance vs. Coverage Ratio and Studies on the Buffer Size}, $\mathrm{f}_{topo} (\cdot)$ is chosen as strategy S1  (Section~\ref{sec:Instantiations of PDGNNs}). 

\subsection{Studies on the Buffer Size $\&$ Performance vs. Coverage Ratio (RQ1, 2, and 3)}\label{sec:Performance vs. Coverage Ratio and Studies on the Buffer Size}
In Table~\ref{tab:performance and coverage ratio}, based on PDGNNs, we compare the proposed \textit{coverage maximization sampling} with uniform sampling and mean of feature (MoF) sampling in terms of coverage ratios and performance when the buffer size (ratio of the dataset) varies from 0.0002 to 0.4 on the OGB-Arxiv dataset.  Our proposed \textit{coverage maximization sampling} achieves a superior coverage ratio, \textcolor{black}{which indeed enhances the performance when the memory budget is tight. In real-world applications, a tight memory budget is a very common situation, making the coverage maximization sampling a favorable choice.} We also notice that the average accuracy for \textit{coverage maximization sampling} is positively related to the coverage ratio in general, which is consistent with the Theorem \ref{thm:equivalence}. 

Table~\ref{tab:performance and coverage ratio} also demonstrates the high memory efficiency of TEM. No matter which sampling strategy is used, the performance can reach $\approx$50\% average accuracy (AA) with only 5\% data buffered. 
In Section~\ref{sec:memory_consumption}, we provide the comparison of the space consumption of different memory based strategies to demonstrate the efficiency of PDGNNs-TEM. 

\begin{figure*}[]
    \centering
    \begin{minipage}{1\textwidth}
        \centering
        \includegraphics[width=1.\textwidth]{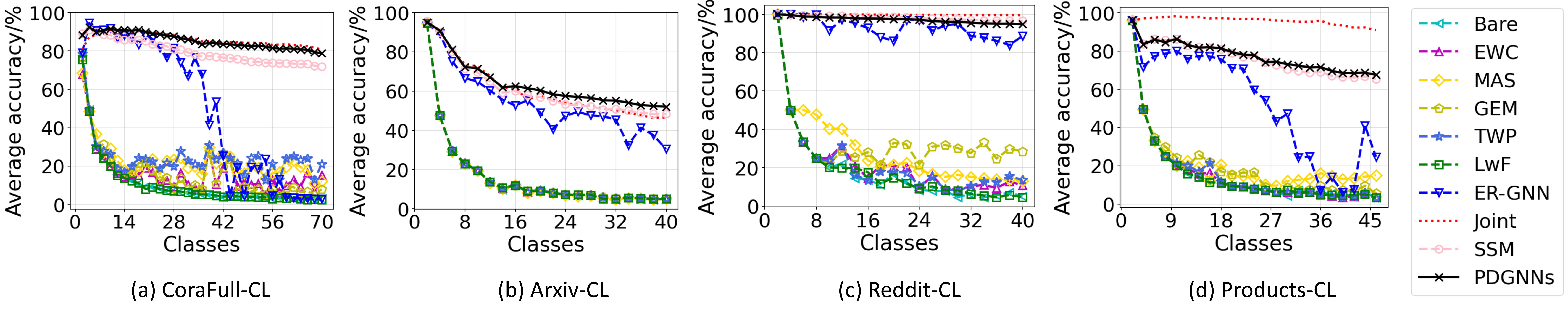}
    \end{minipage}
    \caption{Dynamics of average accuracy in the class-IL scenario.(a) CoraFull, 2 classes per task, 35 tasks. (b) OGB-Arxiv, 2 classes per task, 20 tasks. (c) Reddit, 2 classes per task, 20 tasks. (d) OGB-Products, 2 classes per task, 23 tasks. }
    \label{fig:class-IL corafull_reddit_products}
    \centering
    \begin{minipage}{0.22\textwidth}
        \centering
        \includegraphics[width=1.\textwidth]{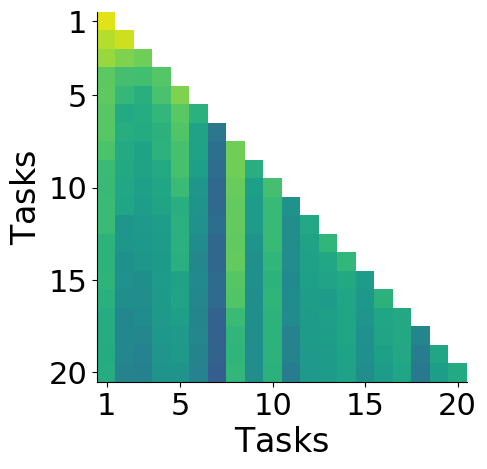}
    \end{minipage}
    \centering
    \begin{minipage}{0.22\textwidth}
        \centering
        \includegraphics[width=1.\textwidth]{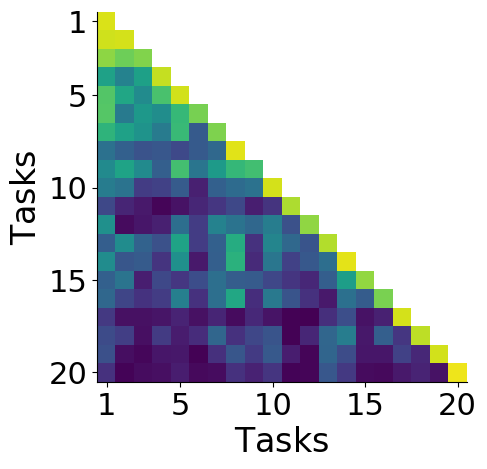}
    \end{minipage}
    \centering
    \begin{minipage}{0.22\textwidth}
        \centering
        \includegraphics[width=1.\textwidth]{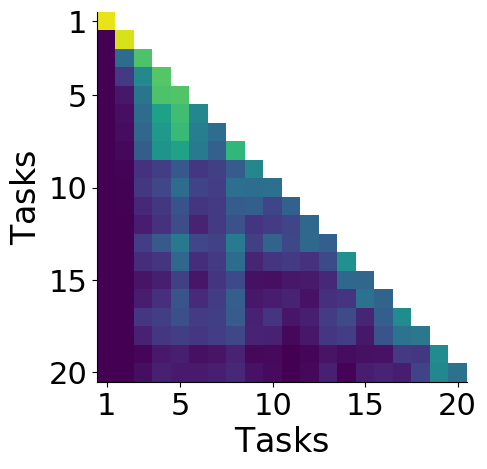}
    \end{minipage}
    \begin{minipage}{0.25\textwidth}
        \centering
        \includegraphics[width=1.\textwidth]{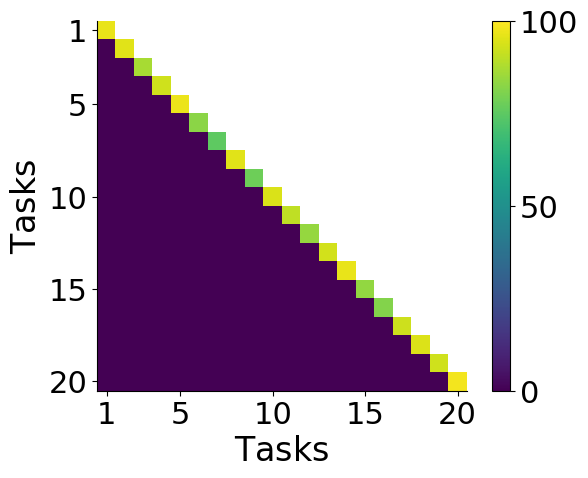}
    \end{minipage}
    \caption{From left to right: accuracy matrix of PDGNNs, ER-GNN, LwF, and Fine-tune on OGB-Arxiv dataset.}
    \label{fig:accmat}
\end{figure*}

\begin{table}[]
\centering
\small
    \caption{Additional space consumption of different memory-replay techniques}
    \begin{tabular}{c|cccc}
    \toprule
      C.L.T.   & CoraFull & OGB-Arxiv  & Reddit & OGB-Products \\ \midrule
        Full Subnetwork & 7,264M & 35M & 2,184,957M & 5,341M \\
        GEM~\cite{lopez2017gradient} &7,840M & 86M& 329M& 82M \\
      ER-GNN~\cite{zhou2021overcoming} &61M &2M &12M &3M  \\
     SSM~\cite{zhang2022sparsified} & 732M& 41M&193M &37M  \\
     SEM~\cite{zhang2023ricci} & 732M& 41M&193M &37M  \\
    PDGNNs-TEM &37M & 2M& 9M&2M \\ \bottomrule
    \end{tabular}
    
    \label{tab:memory consumption}
\end{table}

\subsection{Class-IL and Task-IL Scenarios (RQ4)}
\label{sec:Class-IL and Task-IL}
\textbf{Class-IL Scenario}. As shown in Table~\ref{tab:comparisons with baselines class-IL}, under the class-IL scenario, PDGNNs significantly outperform the baselines and are even comparable to joint training on all 4 public datasets. 
\begin{table*}[]
    \centering
    \caption{Performance comparisons under task-IL on different datasets ($\uparrow$ higher means better).}  
    \begin{tabular}{c||cc|cc|cc|cc}
    \toprule
    \multirow{2}{2.5em}{\textbf{C.L.T.}}  &  \multicolumn{2}{c|}{CoraFull} &  \multicolumn{2}{c|}{OGB-Arxiv} &  \multicolumn{2}{c|}{Reddit} & \multicolumn{2}{c}{OGB-Products}\\ \cline{2-9}
                 & AA/\% $\uparrow$     & AF/\% $\uparrow$  & AA/\% $\uparrow$    & AF /\% $\uparrow$    & AA/\% $\uparrow$    & AF /\%  $\uparrow$    & AA/\%  $\uparrow$    & AF /\% $\uparrow$ \\ \bottomrule\toprule
       Fine-tune &58.0$\pm$1.7&-38.4$\pm$1.8 & 61.7$\pm$3.8&-28.2$\pm$3.3 &73.6$\pm$3.5&-26.9$\pm$3.5 &67.6$\pm$1.6&-25.4$\pm$1.6\\
       EWC~(\citeyear{kirkpatrick2017overcoming}) &78.9$\pm$2.4&-15.5$\pm$2.3 &78.8$\pm$2.7&-5.0$\pm$3.1 &91.5$\pm$4.2&-8.1$\pm$4.6 &90.1$\pm$0.3&-0.7$\pm$0.3\\
       MAS~(\citeyear{aljundi2018memory}) &93.0$\pm$0.5&-0.6$\pm$0.2 &88.4$\pm$0.2&-0.0$\pm$0.0 &98.6$\pm$0.5&-0.1$\pm$0.1 &91.2$\pm$0.6&-0.5$\pm$0.2\\
       GEM~(\citeyear{lopez2017gradient}) &91.6$\pm$0.6&-1.8$\pm$0.6 &87.3$\pm$0.6&2.8$\pm$0.3 &91.6$\pm$5.6&-8.1$\pm$5.8 &87.8$\pm$0.5&-2.9$\pm$0.5\\
       TWP~(\citeyear{liu2021overcoming}) &92.2$\pm$0.5&-0.9$\pm$0.3 &86.0$\pm$0.8&-2.8$\pm$0.8 &87.4$\pm$3.8&-12.6$\pm$4.0 &90.3$\pm$0.1&-0.5$\pm$0.1\\
       LwF~(\citeyear{li2017learning}) &56.1$\pm$2.0&-37.5$\pm$1.8 &84.2$\pm$0.5&-3.7$\pm$0.6 &80.9$\pm$4.3&-19.1$\pm$4.6 &66.5$\pm$2.2&-26.3$\pm$2.3\\
       ER-GNN~(\citeyear{zhou2021overcoming}) &90.6$\pm$0.1&-3.7$\pm$0.1 &86.7$\pm$0.1&11.4$\pm$0.9 &98.9$\pm$0.0&-0.1$\pm$0.1 &89.0$\pm$0.4&-2.5$\pm$0.3\\
       SSM~(\citeyear{zhang2022sparsified}) &\underline{95.8$\pm$0.3}&0.6$\pm$0.2 & 88.4$\pm$0.3&-1.1$\pm$0.1 & \underline{99.3$\pm$0.0}&-0.2$\pm$0.0& \underline{93.2$\pm$0.7}&-1.9$\pm$0.0\\
       SEM-curvature~(\citeyear{zhang2023ricci})  & \textbf{95.9$\pm$0.5}&0.7$\pm$0.4 & \textbf{89.9$\pm$0.3}&-0.1$\pm$0.5 & \textbf{99.3$\pm$0.0}&-0.2$\pm$0.0& \underline{93.2$\pm$0.7}&-1.8$\pm$0.4 \\
       \midrule
       Joint &95.2$\pm$0.2&- & 90.3$\pm$0.2&- &99.4$\pm$0.1&- &91.8$\pm$0.2&-\\ \midrule 
       \textbf{PDGNNs} &{94.6$\pm$0.1}&{0.6$\pm$1.0} &\underline{89.8$\pm$0.4}&{-0.0$\pm$0.5}&{98.9$\pm$0.0}&{-0.5$\pm$0.0} &\textbf{93.5$\pm$0.5}&{-2.1$\pm$0.1}\\
    \bottomrule
    \end{tabular}
    \label{tab:comparisons with baselines task-IL}
\end{table*}
The learning dynamics are shown in Figure~\ref{fig:class-IL corafull_reddit_products}. 
Since the curve of PDGNNs is very close to that of joint training, we conclude that the forgetting problem is nearly eliminated by PDGNNs.
In Table~\ref{tab:comparisons with baselines class-IL} and Figure~\ref{fig:class-IL corafull_reddit_products}, PDGNNs sometimes outperform joint training. The reasons are two-fold. First, PDGNNs learn the tasks sequentially while joint training optimizes the model for all tasks simultaneously, resulting in different optimization difficulties~\citep{bhat2021cilea}. Second, when learning new tasks, joint training accesses all previous data that may be noisy, while replaying the representative TEs may help filter out noise. 
To thoroughly understand different methods, we visualize the accuracy matrices of 4 representative methods, including PDGNNs (memory replay with topological information), ER-GNN (memory replay without topological information), LwF (relatively satisfying performance without memory buffer), and Fine-tune (without continual learning technique), in Figure~\ref{fig:accmat}. 
Compared to the baselines, PDGNNs maintain stable performance on each task even though new tasks are continuously learned. 

\noindent\textbf{Task-IL Scenario.} The comparison results under the task-IL scenario are shown in Table~\ref{tab:comparisons with baselines task-IL}. We can observe that PDGNNs still outperform most baselines on all different datasets and is comparable to SEM \cite{zhang2023ricci} and SSM \cite{zhang2022sparsified}, even though task-IL is less challenging than the class-IL as we discussed in Section~\ref{sec:experiment setup}. 

\begin{figure*}[h]
    \centering
    \includegraphics[width=0.82\textwidth]{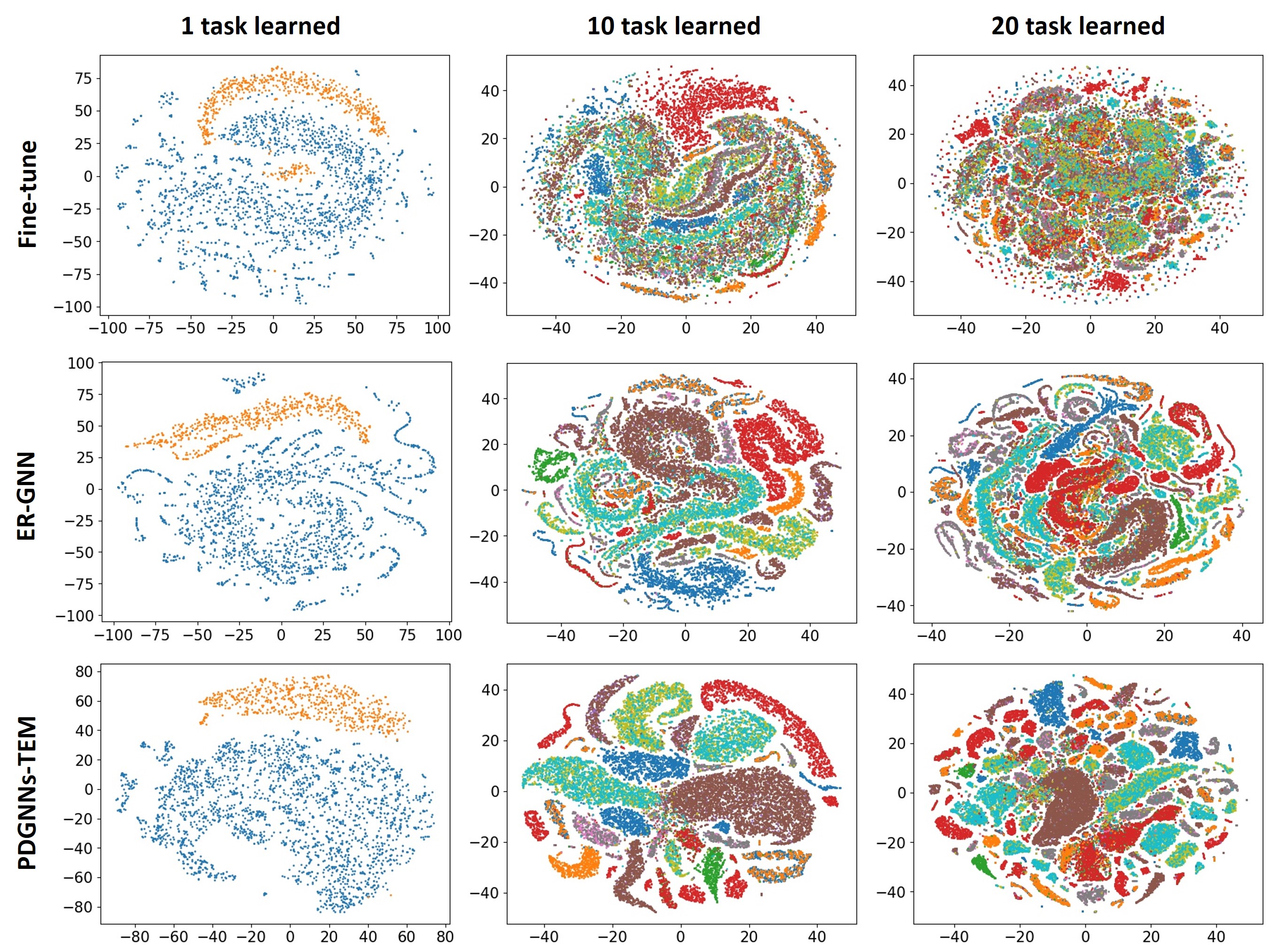}
    \caption{Visualization of the node embeddings of different classes of Reddit, after learning 1, 10, and 20 tasks. From the top to the bottom, we show the results of Fine-tune, ER-GNN, and PDGNNs-TEM. Each color corresponds to a class.}
    \label{fig:tsne}
\end{figure*}

\subsection{Memory Consumption Comparison (RQ1)}\label{sec:memory_consumption}
Memory-replay based methods outperform other methods, but also consume additional memory space.
In this subsection, we compare the space consumption of different memory designs to demonstrate the memory efficiency of PDGNNs-TEM. 
The final memory consumption (measured by the number of float32 values) after learning each entire dataset is shown in Table~\ref{tab:memory consumption}. As a reference, the memory consumption of storing full computation ego-subnetwork is also calculated.
According to Table~\ref{tab:memory consumption}, storing full subnetworks costs intractable memory usage on dense networks like Reddit, and the strategy to buffer gradients also incurs high memory cost (GEM). SSM could significantly reduce memory consumption with the sparsification strategy. Both PDGNNs-TEM and ER-GNN are highly efficient in terms of memory space usage. While PDGNNs-TEM exhibits superior performance compared to ER-GNN. 

\subsection{Interpretation of Node Embeddings (RQ5)}\label{sec:tsne}

To interpret the learning process of PDGNNs-TEM, we visualize the node embeddings of different classes with t-SNE~\citep{van2008visualizing} while learning on a task sequence of 20 tasks over the Reddit dataset. In Figure~\ref{fig:tsne}, besides PDGNNs-TEM that replay data with topological information, we also include two representative baselines for comparison, \textit{i.e.}, ER-GNN to show how the lack of topological information may affect the node embeddings, and Fine-tune to show the results without any continual learning technique. As shown in Figure~\ref{fig:tsne}, PDGNNs-TEM can well separate the nodes from different classes even when node types of nodes are continuously been involved (in new tasks). In contrast, for ER-GNN and Fine-tune, the boundaries of different classes are less clear, especially when more tasks are continuously learned.

\color{black}
\section{Conclusion}
In this work, we propose a general framework of Parameter Decoupled Graph Neural Networks (PDGNNs) with Topology-aware Embedding Memory (TEM) for continual learning on expanding networks. Based on the Topology-aware Embeddings (TEs), we reduce the space complexity of the memory buffer from $\mathcal{O}(nd^L)$ to $\mathcal{O}(n)$, which enables PDGNNs to fully utilize the explicit topological information sampled from the previous tasks for retraining. We also discover and theoretically analyze the pseudo-training effect of TEs. The theoretical findings inspire us to develop the \textit{coverage maximization sampling} strategy, which has been demonstrated to be \textcolor{black}{highly efficient when the memory budget is tight.} Finally, thorough empirical studies, including comparison with the state-of-the-art methods in both class-IL and task-IL continual learning scenarios, demonstrate the effectiveness of PDGNNs with TEM.

\section{Acknowledgement}
 Dongjin Song was supported by the National Science Foundation (NSF) Grant No. 2338878. Yixin Chen is supported by NSF grant CBE-2225809 and DOE grant DE-SC0024702.

\clearpage
\bibliographystyle{ACM-Reference-Format}
\balance
\bibliography{sample-base}

\appendix
\newtheorem{definition_app}{Definition}
\newtheorem{thm_app}{Theorem}
\newtheorem{lemma_app}{Lemma}

\section{Theoretical Analysis}\label{sec:additonal theoretical results}
In this section, we give proofs and analysis of the theoretical results.

\begin{proof}[\textcolor{black}{Proof of Theorem \ref{thm:equivalence}.1}]

Given a node $v$, the prediction is:
\begin{align}
    \mathbf{\hat{y}}_v = \mathrm{f}_{out}(\mathbf{e}_v; \boldsymbol{\theta})
\end{align}
$\because \mathbf{e}_v= \sum_{w \in \mathbb{V}_v^{sub}} \mathbf{x}_w \cdot \pi(v, w; \hat{\mathbf{A}})$, where $\mathbb{V}_v^{sub}$ denotes the node set of the computation ego-subnetwork $\mathcal{G}^{sub}_v$, and $\hat{\mathbf{A}}$ is the adjacency matrix of $\mathcal{G}^{sub}_v$. 

$\therefore$ 
\begin{align}
    \mathbf{\hat{y}}_v = \mathrm{f}_{out}\Big(\sum_{w \in \mathbb{V}_v^{sub}} \mathbf{x}_w \cdot \pi(v, w; \hat{\mathbf{A}}); \boldsymbol{\theta}\Big)
\end{align}
Given the label of node $v$ ($y_v$), the objective function of training the model with node $v$ is formulated as:
\begin{align}\label{eq:lossv}
    \mathcal{L}_v = l\Bigg(\mathrm{f}_{out}\Big(\sum_{w \in \mathbb{V}_v^{sub}} \mathbf{x}_w \cdot \pi(v, w; \hat{\mathbf{A}}); \boldsymbol{\theta}\Big), \mathbf{y}_v \Bigg) ,
\end{align}
where $l$ could be any loss function.
Since $\mathbb{V}_v^{sub}$ contains both the features of node $v$ and its neighbors, Equation \ref{eq:lossv} can be further expanded to separate the contribution of node $v$ and its neighbors: 
\begin{align}\label{eq:lossv_expand}
    \mathcal{L}_v = l\Bigg(\mathrm{f}_{out}\Big(\underbrace{\mathbf{x}_v \cdot \pi(v,v;\hat{\mathbf{A}})}_{\text{information from node $v$}}+\underbrace{\sum_{w \in \mathbb{V}_v^{sub}\backslash \{v\}} \mathbf{x}_w \cdot \pi(v, w; \hat{\mathbf{A}})}_{\text{neighborhood information}}; \boldsymbol{\theta}\Big), \mathbf{y}_v \Bigg) ,
\end{align}
Given an arbitrary node $q\in \mathbb{V}_v^{sub}$ but $q\neq v\in\mathbb{V}_v^{sub}$ (the adjacency matrix $\hat{\mathbf{A}}$ stays the same), we can similarly obtain the loss of training the model with node $q$:
\begin{align}\label{eq:lossq_expand}
     \mathcal{L}_q = l\Bigg(\mathrm{f}_{out}\Big(\underbrace{\mathbf{x}_q \cdot \pi(q,q;\hat{\mathbf{A}})}_{\text{information from node $q$}}+\underbrace{\sum_{w \in \mathbb{V}_q^{sub}\backslash \{q\}} \mathbf{x}_w \cdot \pi(q, w; \hat{\mathbf{A}})}_{\text{neighborhood information}}; \boldsymbol{\theta}\Big), \mathbf{y}_q \Bigg) .
\end{align}

Since $q \in \mathbb{V}_v^{sub}\backslash \{v\}$, we rewrite \textcolor{black}{Equation \ref{eq:lossv_expand}} as:
\begin{align}\label{eq:lossv_expand_new}
     \mathcal{L}_v = l\Bigg(\mathrm{f}_{out}\Big(\underbrace{\mathbf{x}_q \cdot \pi(v,q;\hat{\mathbf{A}})}_{\text{information from node $q$}}+\underbrace{\sum_{w \in \mathbb{V}_v^{sub}\backslash \{q\}} \mathbf{x}_w \cdot \pi(v, w; \hat{\mathbf{A}})}_{\text{neighborhood information}}; \boldsymbol{\theta}\Big), \textbf{y}_v \Bigg) ,
\end{align}
By comparing Equation \ref{eq:lossv_expand_new} and \ref{eq:lossq_expand}, we could observe the similarity in the loss of node $v$ and $q$, and the difference lies in the contribution (weight $\pi(\cdot, \cdot; \hat{\mathbf{A}})$) of each node and the neighboring nodes ($\mathbb{V}_q^{sub}$ and $\mathbb{V}_v^{sub}$). 
\end{proof}

\begin{proof}[\textcolor{black}{Proof of Theorem \ref{thm:equivalence}.2}]
In this part, we choose the loss function $l$ as cross entropy $\mathrm{CE}(\cdot,\cdot)$, which is the common choice for classification problems. In the following, we will first derive the gradient of training the PDGNNs with ($\mathbf{e}_v$, $y_v$). For cross entropy, we denote the one-hot vector form label as $\mathbf{y}_v$, of which the $y_v$-th element is one and other entries are zero.
Given the loss of a node $v$ as shown in the Equation \ref{eq:lossv}, the gradient is derived as:
\begin{align}\label{eq:lossv_rewrite}
    \nabla_{\boldsymbol{\theta}}\mathcal{L}_v &= \nabla_{\boldsymbol{\theta}} \mathrm{CE}\Bigg(\sum_{w \in \mathbb{V}_v^{sub}} \mathrm{f}_{out}\Big(\mathbf{x}_w \cdot \pi(v, w; \hat{\mathbf{A}}); \boldsymbol{\theta}\Big), \mathbf{y}_v \Bigg) &&\\
    &= \nabla_{\boldsymbol{\theta}} \Bigg(\mathbf{y}_{v,k} \cdot \log \sum_{w \in \mathbb{V}_v^{sub}} \mathrm{f}_{out}\Big(\mathbf{x}_w \cdot \pi(v, w; \hat{\mathbf{A}}); \boldsymbol{\theta}\Big)_{k}  \Bigg) &&\\
    &= \mathbf{y}_{v,k} \cdot \frac{\nabla_{\boldsymbol{\theta}}\Big(\sum_{w \in \mathbb{V}_v^{sub}} \mathrm{f}_{out}\big(\mathbf{x}_w \cdot \pi(v, w; \hat{\mathbf{A}}); \boldsymbol{\theta}\big)_{k}\Big)}{ \sum_{w \in \mathbb{V}_v^{sub}} \mathrm{f}_{out}\Big(\mathbf{x}_w \cdot \pi(v, w; \hat{\mathbf{A}}); \boldsymbol{\theta}\Big)_{k}}&& \\
    &= \mathbf{y}_{v,k} \cdot \frac{\sum_{w \in \mathbb{V}_v^{sub}} \nabla_{\boldsymbol{\theta}}\mathrm{f}_{out}\Big(\mathbf{x}_w \cdot \pi(v, w; \hat{\mathbf{A}}); \boldsymbol{\theta}\Big)_{k}}{ \sum_{w \in \mathbb{V}_v^{sub}} \mathrm{f}_{out}\Big(\mathbf{x}_w \cdot \pi(v, w; \hat{\mathbf{A}}); \boldsymbol{\theta}\Big)_{k}}&& \\
    &= \mathbf{y}_{v,k}\cdot \frac{\sum_{w \in \mathbb{V}_v^{sub}} \nabla_{\boldsymbol{\theta}}\mathrm{f}_{out}(\mathbf{x}_w; \boldsymbol{\theta})_{k} \cdot \pi(v, w; \hat{\mathbf{A}})}{ \sum_{w \in \mathbb{V}_v^{sub}} \mathrm{f}_{out}\Big(\mathbf{x}_w \cdot \pi(v, w; \hat{\mathbf{A}}); \boldsymbol{\theta}\Big)_{k}}&& \\
    &=  \frac{\sum_{w \in \mathbb{V}_v^{sub}} \mathbf{y}_{v,k} \cdot\frac{\nabla_{\boldsymbol{\theta}}\mathrm{f}_{out}(\mathbf{x}_w; \boldsymbol{\theta})_{k}}{\mathrm{f}_{out}(\mathbf{x}_w; \boldsymbol{\theta})_{k}} \cdot \mathrm{f}_{out}(\mathbf{x}_w; \boldsymbol{\theta})_{k} \cdot \pi(v, w; \hat{\mathbf{A}})}{ \sum_{w \in \mathbb{V}_v^{sub}} \mathrm{f}_{out}\Big(\mathbf{x}_w \cdot \pi(v, w; \hat{\mathbf{A}}); \boldsymbol{\theta}\Big)_{k}} &&\\
    &= \frac{\sum_{w \in \mathbb{V}_v^{sub}} \nabla_{\boldsymbol{\theta}} \mathrm{CE}\big(\mathrm{f}_{out}(\mathbf{x}_w; \boldsymbol{\theta}),\mathbf{y}_{v,k} \big) \cdot \mathrm{f}_{out}(\mathbf{x}_w; \boldsymbol{\theta})\cdot \pi(v, w; \hat{\mathbf{A}})}{ \sum_{w \in \mathbb{V}_v^{sub}} \mathrm{f}_{out}\Big(\mathbf{x}_w \cdot \pi(v, w; \hat{\mathbf{A}}); \boldsymbol{\theta}\Big)} &&\\
    &= \sum_{w \in \mathbb{V}_v^{sub}} \frac{\mathrm{f}_{out}(\mathbf{x}_w; \boldsymbol{\theta})\cdot\pi(v, w; \hat{\mathbf{A}})}{ \sum_{w \in \mathbb{V}_v^{sub}} \mathrm{f}_{out}\Big(\mathbf{x}_w \cdot \pi(v, w; \hat{\mathbf{A}}); \boldsymbol{\theta}\Big)}\cdot && \\ & \quad\quad \nabla_{\boldsymbol{\theta}} \mathrm{CE}\big(\mathrm{f}_{out}(\mathbf{x}_w; \boldsymbol{\theta}),\mathbf{y}_{v} \big).&&\label{eq:15}
\end{align}
The loss of training $\mathrm{f}_{out}(\mathbf{x}_w; \boldsymbol{\theta})$ with pairs of feature and pseudo-label ($\mathbf{x}_w$, $y_v$) of all nodes of $\mathcal{G}^{sub}_v$ is:
\begin{align}
    \mathcal{L}_{\mathcal{G}^{sub}_v} = \sum_{w \in \mathbb{V}_v^{sub}} \mathrm{CE}\big(\mathrm{f}_{out}(\mathbf{x}_w; \boldsymbol{\theta}), \mathbf{y}_{v} \big) \\
\end{align}
Then, the corresponding gradient of $\mathcal{L}_{\mathcal{G}^{sub}_v}$ is :
\begin{align}
    \nabla_{\boldsymbol{\theta}}\mathcal{L}_{\mathcal{G}^{sub}_v} = \sum_{w \in \mathbb{V}_v^{sub}} \nabla_{\boldsymbol{\theta}}\mathrm{CE}\big(\mathrm{f}_{out}(\mathbf{x}_w; \boldsymbol{\theta}), \mathbf{y}_{v} \big).\label{eq:16}
\end{align}
By comparing Equation \ref{eq:15} and \ref{eq:16}, we can see that training PDGNNs with a topology-aware embedding $\mathbf{e}_v$ equals to training the function $\mathrm{f}_{out}(\cdot;\boldsymbol\theta)$ on all nodes of the computation ego-subnetwork $\mathcal{G}^{sub}_v$ with a weight $\frac{\mathrm{f}_{out}(\mathbf{x}_w; \boldsymbol{\theta})\cdot\pi(v, w; \hat{\mathbf{A}})}{ \sum_{w \in \mathbb{V}_v^{sub}} \mathrm{f}_{out}\big(\mathbf{x}_w \cdot \pi(v, w; \hat{\mathbf{A}}); \boldsymbol{\theta}\big)}$ on each node to rescale the contribution dynamically.
\end{proof}

\clearpage

\end{document}